\documentclass{article}

\usepackage[preprint]{neurips_2026}

\usepackage[utf8]{inputenc} 
\usepackage[T1]{fontenc}    
\usepackage{hyperref}       
\usepackage{url}            
\usepackage{booktabs}       
\usepackage{amsfonts}       
\usepackage{nicefrac}       
\usepackage{microtype}      
\usepackage{xcolor}         

\usepackage{amsmath}
\usepackage{amssymb}
\usepackage{mathtools}
\usepackage{amsthm}
\usepackage{graphicx}
\usepackage{array}
\usepackage{algorithm}
\usepackage{algorithmic}
\usepackage[capitalize,noabbrev]{cleveref}

\usepackage{wrapfig}

\title{MAG: MAnifold Guided Semi-Supervised Multi-modal In-Context Learning}

\author{
\normalfont
\textbf{Zirui Cheng}$^{1,2}$,
\textbf{Xun Xu}$^{2}$,
\textbf{Tiankai Chen}$^{2}$,
\textbf{Fady Rezk}$^{2}$,
\textbf{Bowen Zheng}$^{1}$, \\
\textbf{Xiaodong Shi}$^{1}$,
\textbf{Shijie Li}$^{2}$,
\textbf{Kangkang Lu}$^{2}$,
\textbf{Bharadwaj Veeravalli}$^{1}$,
\textbf{Nancy F. Chen}$^{2}$ \\[3pt]
$^{1}$College of Design and Engineering, National University of Singapore \\
$^{2}$Institute for Infocomm Research, A\*STAR, Singapore \\
\texttt{zirui\_c@u.nus.edu}
}

\begin{document}

\maketitle

\begin{abstract}
Few-shot in-context learning (ICL) with multi-modal large language models (MLLMs) enables task adaptation without parameter updates, but its performance is highly sensitive to the quality and coverage of the selected demonstrations. While unlabeled multi-modal data is abundant, it remains elusive how to exploit them for ICL. We propose \textbf{MAG} (MAnifold-Guided semi-supervised in-context demonstration selection), an efficient framework that leverages unlabeled data to improve multi-modal ICL. MAG formulates demonstration selection as a semi-supervised propagation problem on a multi-modal graph and adopts a two-stage strategy: (i) relevance score propagation identifies a compact set of high-impact unlabeled samples for pseudo-labeling, reducing MLLM inference cost; (ii) multi-modal relevance is used to select the final demonstrations. We show that textual representations are more effective for relevance propagation, while both visual and textual modalities are crucial for high-quality demonstration selection. Experiments on eight multi-modal benchmarks demonstrate that MAG consistently outperforms strong baselines in label-scarce regimes, achieving significant gains with a limited pseudo-labeling budget.
\end{abstract}

\section{Introduction}

Large language models (LLMs) and their multi-modal extensions (MLLMs) have shown strong {in-context learning} (ICL) abilities, allowing adaptation to new tasks by conditioning on a small set of input–output demonstrations, e.g., visual samples, textual queries, and answer tuples, without parameter updates~\citep{brown2020language,alayrac2022flamingo,liu2023llava,openai2023gpt4}.
In multi-modal settings, ICL is especially attractive, as it enables MLLMs to address diverse vision–language tasks through prompt construction alone.
However, the effectiveness of few-shot multi-modal ICL critically depends on the quality, relevance, and coverage of the demonstrations, making their construction both central and challenging~\citep{zhou2024visual,wu2023self,chen2025can,chen2025provoking}.


In practice, such high-quality input–output demonstrations, i.e., labeled data, are often scarce. For instance, new users may have limited historical interactions, or insufficient labeled data may be available to bootstrap a new downstream task. This inherent label scarcity substantially limits the effectiveness of ICL~\citep{levy2023diverse,rahman2025mpcar}.
In contrast, it is typically much easier to acquire large amounts of unlabeled multi-modal data, such as visual samples paired with textual queries but without answers. Yet, in the absence of ground-truth outputs, these unlabeled samples are difficult to exploit using standard retrieval-based ICL methods~\citep{wu2023self,rubin2022learning}. 
This tension naturally motivates semi-supervised learning which has long shown that unlabeled data can significantly improve generalization by capturing intrinsic data structure that is inaccessible from limited labeled examples~\citep{zhu2003semi,tarvainen2017mean,iscen2019label,sohn2020fixmatch}. These insights suggest that abundant unlabeled multi-modal data, if properly utilized, could serve as a powerful resource for constructing more informative and robust in-context demonstrations.\begin{wrapfigure}{r}{0.55\textwidth}
    \centering
\includegraphics[width=\linewidth]{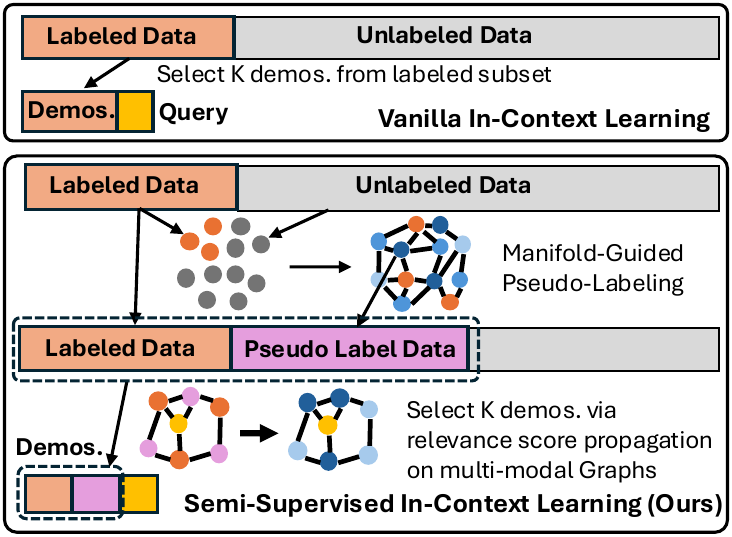}
\vspace{-0.5cm}
    \caption{Comparison of our proposed semi-supervised in-context learning (ICL) against vanilla ICL.}
    \label{fig:teaser}
    \vspace{-0.8cm} 
\end{wrapfigure}

A central challenge is how to effectively propagate information from labeled data to unlabeled data in a manner compatible with ICL. We address this challenge through a two-stage framework designed to balance scalability with demonstration quality.

First of all, given the pool of unlabeled multi-modal samples and the high inference cost of MLLMs, directly pseudo-labeling all unlabeled data is too expensive. Instead, we  employ a graph-based relevance score propagation mechanism~\citep{zhu2002learning,zhou2004learning} to identify a compact candidate set of unlabeled samples that are most relevant to the labeled data. This step propagates supervision along the intrinsic \textbf{data manifold}, enabling efficient filtering of unlabeled samples that are likely to provide informative and transferable context. Importantly, the procedure is formulated as label propagation on the graph which is both highly computational efficient with closed-form solution and offers principled mathematical explanation. The approach is also highly scalable to very large unlabeled data pool due to the sparsity of the manifold graph. Eventually, only selected unlabeled candidates are then forwarded to the MLLM for pseudo-labeling, reducing inference cost while preserving semantic relevance.
In the second stage, we further select the most effective in-context demonstrations from the combined labeled and pseudo-labeled candidate pool. Specifically, we apply score propagation again by treating the query sample as positive node and propagate relevance score to all candidates for demonstration selection.

Critically, we reveal that relying on textual descriptions of visual content is a stable option for relevance propagation in the first stage for coarse filtering. In contrast, incorporating both visual and textual modalities is essential for high-quality demonstration selection in the second stage to ensure fine-grained alignment between the demonstrations and the query. In particular, late fusion between visual and textual modalities proves more effective in the second stage, as it better synergizes complementary modalities while decoupling modality-specific features.

To reflect the \textbf{MA}nifold-\textbf{G}uided nature of our semi-supervised in-context demonstration selection, we term the proposed method \textbf{MAG}. Unlike learning-based approaches~\citep{deng2024enhancing,li2025taco} that rely on fine-tuning or training auxiliary models, MAG is learning-free, enabling rapid deployment across diverse tasks.
Across eight benchmarks spanning visual emotion recognition~\citep{yang2023emoset,panda2018contemplating}, scene text understanding~\citep{zong2024vlicl}, visual reasoning~\citep{chen2024we}, and visual question answering~\citep{hudson2019gqa,okvqa}, MAG consistently outperforms methods retrieving from labeled data alone. 
These results demonstrate that principled incorporation of unlabeled data through relevance-guided pseudo-labeling and multi-modal graph propagation substantially improves ICL performance under label-scarce conditions, with particularly pronounced gains on tasks requiring complex cross-modal reasoning.

We summarize the contributions as follows.
\begin{itemize}
    \item 
    We identify label scarcity as a key bottleneck in few-shot multi-modal ICL and propose to leverage abundant unlabeled multi-modal data via a semi-supervised formulation, enabling more robust and informative demonstration construction under limited labeled data.

\item 
We introduce a two-stage, training-free framework that employs graph-based relevance score propagation to efficiently select a compact, query-relevant subset of unlabeled samples for pseudo-labeling and ICL, substantially reducing MLLM inference cost while preserving semantic relevance.

\item Across 8 diverse benchmarks, 
      we show consistent and substantial gains with particularly large improvements on reasoning-intensive tasks, demonstrating that unlabeled data can be 
      effectively leveraged for MAG.
\end{itemize}

\section{Related Work}

\vspace{-0.3cm}
\noindent\textbf{In-Context Learning (ICL)}: ICL enables large language models to solve new tasks by conditioning on demonstrations without parameter updates~\citep{brown2020language}. Research has explored ICL mechanisms as implicit gradient descent~\citep{peebles2022learning,garg2022can} or Bayesian inference~\citep{xie2022explanation}, while other work examines how demonstration selection, ordering, and formatting affect performance~\citep{lu2022fantastically,rubin2022learning,an2023context}.
Methods for improving ICL fall into learning-free and learning-based categories. 
Learning-based approaches optimize prompts or selection strategies through training. Examples include automatic prompt engineering~\citep{zhou2022large}, gradient-based prompt optimization~\citep{pryzant2023automatic}, and parameter-efficient adaptation~\citep{liu2022few}. CEIL~\citep{ye2023compositional} learns selection policies using determinant point processes. While effective, these methods require additional training or task-specific supervision, unlike our learning-free approach.
Learning-free methods use heuristics for demonstration selection: TopK+MDL uses content similarity and minimum description length~\citep{wu2023self}, while Cover-LS emphasizes structural similarity and diversity~\citep{levy2023diverse}. MAPLE~\citep{chen2025maple} introduces graph-based influence propagation to select demonstrations and incorporates unlabeled data through adaptive pseudo-labeling for text-only ICL. Unlike MAPLE, which addresses many-shot pseudo-labeling for in-context learning with large context windows, our work targets few-shot multi-modal ICL by introducing a two-stage, modality-aware score propagation that efficiently filters unlabeled data and selects high-quality demonstrations under strict context length and inference cost constraints. The introduced score propagation presents a more computational efficient and mathematically principled perspective. We also identify effective multi-modal fusion practices in the context of ICL.


\noindent\textbf{Multi-Modal In-Context Learning}:
Vision-language models~\citep{radford2021learning,jia2021scaling,li2022blip} established shared representation spaces, enabling multi-modal ICL with models like Flamingo~\citep{alayrac2022flamingo}, GPT-4V~\citep{openai2023gpt4}, and LLaVA~\citep{liu2023llava}. However, MM-ICL performance remains highly sensitive to demonstration selection~\citep{li2024configure,chen2025can}.
Learning-free MM-ICL methods address this through improved retrieval. MMICES~\citep{chen2025can} and VICL~\citep{zhou2024visual} combine visual concept consistency with textual similarity, while ICCG~\citep{li2024context} uses diversity-coverage matching. MPCAR~\citep{rahman2025mpcar} incorporates multi-perspective demonstrations, and Cola~\citep{chen2023large} coordinates multiple VLMs. However, these methods operate only on labeled demonstrations and cannot leverage abundant unlabeled data, limiting their effectiveness in label-scarce settings.
Learning-based methods adapt MLLMs through training. STIC~\citep{deng2024enhancing} uses self-training for image comprehension, TACO~\citep{li2025taco} employs task-aware attention, CVR-LLM~\citep{li2024enhancing} generates context-aware textual descriptions, and CONTEXTNA~\citep{li2024enhancing} proposes an agentic retrieval framework. These methods require training or complex pipelines, while our approach is learning-free and directly applicable across tasks.

\noindent\textbf{Semi-Supervised Learning and Label Propagation}:
Label propagation diffuses label information from labeled to unlabeled samples over similarity graphs~\citep{zhu2002learning,zhou2004learning}, assuming nearby samples share labels. Modern variants leverage deep representations: Iscen et al.~\citep{iscen2017efficient} proposed efficient diffusion on region manifolds, while subsequent work integrated pseudo-labeling into deep learning~\citep{iscen2019label,basak2023pseudo,zheng2023simmatchv2}. More expressive approaches use hypergraph structures~\citep{zhang2020hypergraph} to model higher-order relations.
Chen et al.~\citep{chen2025cross} propose graph-based pseudo-labeling for text-only ICL, propagating labels over example-query relations. To the best of our knowledge, no prior work has applied label propagation paradigms to multi-modal ICL. We build on these advances by extending influence-guided pseudo-labeling to multi-modal ICL and demonstrating that visual and textual graphs capture complementary structures requiring late fusion.

\section{Methodology}

\vspace{-0.3cm}
\begin{figure*}[t]
\vspace{-0.2cm}
\begin{center}
\includegraphics[width=1\textwidth]{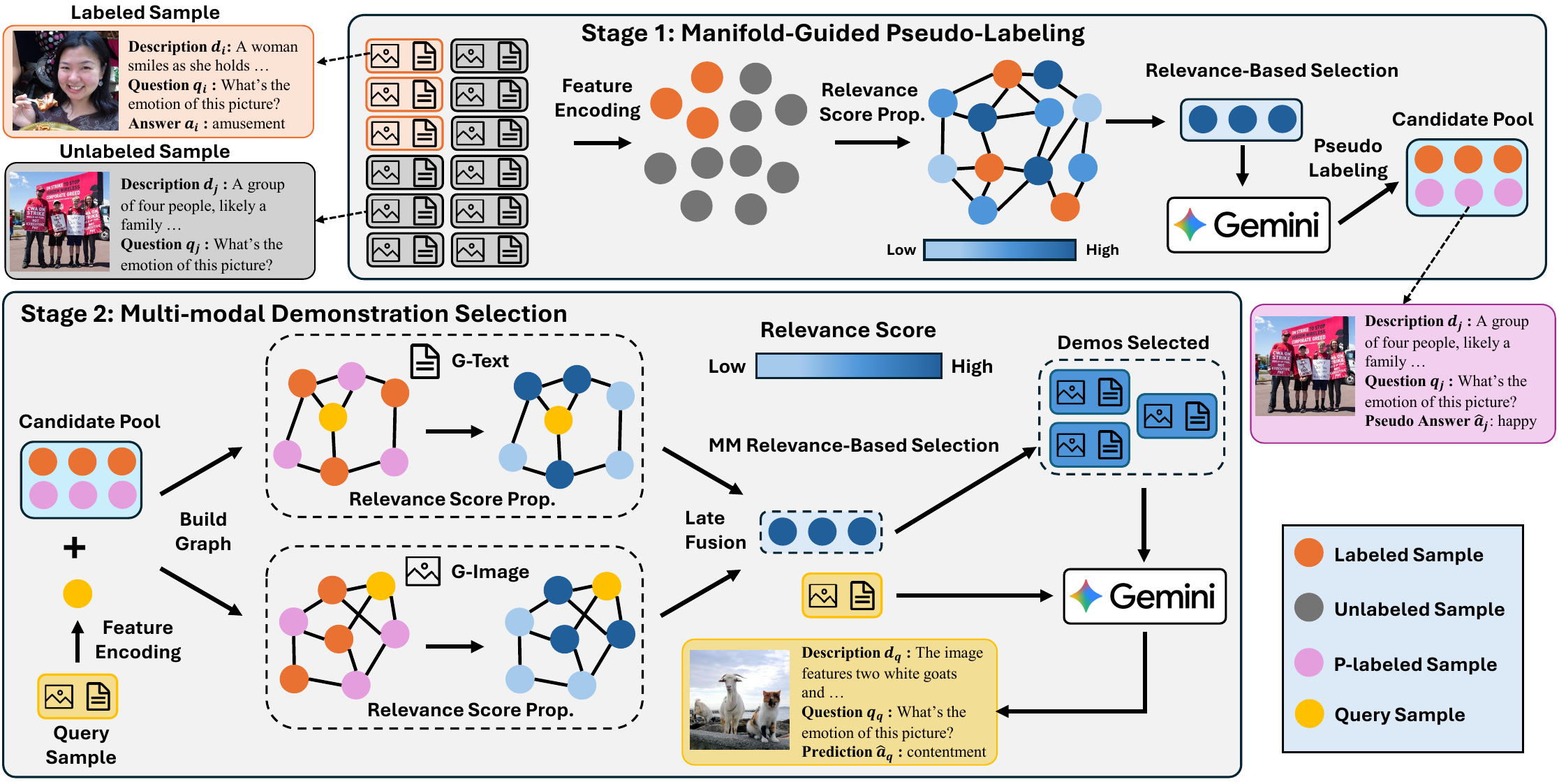}
\vspace{-0.5cm}
\caption{Overview of MAG. Stage~1 (top): We generate textual descriptions of all images using the MLLM, construct a textual relationship graph, identify high-relevance unlabeled samples, and pseudo-label them. Stage~2 (bottom): We construct visual and textual graphs over the expanded candidate pool and perform query-conditioned relevance propagation in both modalities to select demonstrations for each query.}
\label{fig:framework}
\end{center}
\vspace{-0.7cm}
\end{figure*}

\paragraph{Problem Setup.}
We are given a labeled dataset
$\mathcal{D}_L = \{(x_i, q_i, a_i)\}_{i=1}^{N_L}$
and an unlabeled dataset
$\mathcal{D}_U = \{(x_j, q_j)\}_{j=1}^{N_U}$,
where $x_i$, $q_i$, and $a_i$ denote the image, text (e.g., a question in VQA), and answer, respectively.
In typical semi-supervised settings, $N_L \ll N_U$.
Our goal is to select effective in-context demonstrations for each query $q$ under label-scarce conditions.

MAG addresses this problem through two components:
(1) \textbf{manifold-guided pseudo-labeling}, which identifies and pseudo-labels high-relevance unlabeled samples to expand the candidate pool while controlling cost;
and (2) \textbf{query-conditioned multi-modal graph-based demonstration selection}, which performs relevance propagation on complementary visual and textual graphs to select most relevant demonstrations from the candidate pool.

\subsection{Manifold-Guided Pseudo-Labeling}
\vspace{-0.2cm}
\label{subsec:stage1}

In the first stage, we identify unlabeled samples that are strongly connected to labeled data and pseudo-label them to augment the limited labeled pool.
To this end, we construct a textual relationship graph that captures semantic structure shared across labeled and unlabeled samples.

\noindent\textbf{Textual Description Generation.}
Contrary to the practice in MAPLE~\cite{chen2025maple} which directly uses question $q_i$ to compute affinity, we generate textual descriptions of image using the MLLM, $d_i = \mathcal{M}_{\mathrm{desc}}(x_i)$, 
where $\mathcal{M}_{\mathrm{desc}}$ produces a detailed description of the visual content.
These descriptions, combined with any existing textual information (e.g., task-specific questions), form the textual representation $t_i=[d_i,q_i]$. This design ensures more visual information is captured in the construction of graph affinity and enables handling the tasks where questions are identical across samples (e.g. many VQA tasks have an identical question for different images).

\noindent\textbf{Manifold Graph Construction.}
We construct a textual relationship graph
$\mathcal{G}^{(t)} = (\mathcal{V}^{(t)}, \mathcal{E}^{(t)})$
over $\mathcal{D}_L \cup \mathcal{D}_U$.
Each node corresponds to a text embedding
$h_i^{(t)} = f_t(t_i)$,
where $f_t$ is a pre-trained text encoder (e.g., Contriver~\citep{izacard2022unsuperviseddenseinformationretrieval}).
We build a $k$-NN graph and symmetrize it to capture the underlying manifold, with edge weights defined as
\begin{equation}\label{eq:knn_graph}
w_{ij}^{(t)} =
\begin{cases}
s_{ij}^{(t)}, & \text{if } j \in \mathrm{NN}_k(i), \\
0, & \text{otherwise},
\end{cases}
\quad
s_{ij}^{(t)} = \langle h_i^{(t)}, h_j^{(t)} \rangle .
\end{equation}

\noindent\textbf{Relevance Score Propagation.}
To identify candidate unlabeled samples for pseudo-labeling, we propagate relevance scores from labeled nodes to unlabeled ones.
We initialize an relevance score vector
$I^{(t)}_{(0)} \in \{0,1\}^{N_L + N_U}$
by assigning value $1$ to labeled samples and $0$ to unlabeled samples.
Relevance is iteratively propagated over $\mathcal{G}^{(t)}$ via
\begin{equation}
I^{(t)}_{(k)} = \alpha \hat{W}^{(t)} I^{(t)}_{(k-1)} + (1-\alpha) I^{(t)}_{(0)},
\end{equation}
where $k$ indicates the propagation step, $\alpha \in [0,1]$ controls the propagation strength and
$\hat{W}^{(t)} = (D^{(t)})^{-1/2} W^{(t)} (D^{(t)})^{-1/2}$ is the normalized adjacency matrix with $D^{(t)}$ being the degree matrix.
The iteration converges to the closed-form solution
\begin{equation}\label{eq:closedform}
I^{(t)}_{(\infty)} = (1-\alpha)(\mathbf{I} - \alpha \hat{W}^{(t)})^{-1} I^{(t)}_{(0)}.
\end{equation}

We select the top-$K$ unlabeled samples with the highest relevance scores and pseudo-label them using the MLLM conditioned on labeled demonstrations:
\begin{equation}
\hat{a}_j = \mathcal{M}(x_j, q_j; \mathcal{D}_L).
\end{equation}
The pseudo-labeled set $\mathcal{D}_P$ and labeled set $\mathcal{D}_L$ together form the expanded candidate pool
$\mathcal{D}_E = \mathcal{D}_L \cup \mathcal{D}_P$. The above score propagation is easily scalable to very large unlabeled data pool since $W$ is sparse which allows efficiency solution to Eq.~\ref{eq:closedform}, e.g. through conjugate gradient or Cholesky decomposition. 

\subsection{Multi-Modal Graph Construction \& Query-Conditioned Propagation}

In the second stage, we select the most relevant demonstrations for ICL from the expanded candidate pool $\mathcal{D}_E$. 
We construct separate relationship graphs in the visual and textual embedding spaces. These practices ensure most relevant demonstrations are selected for ICL while the inference cost is manageable.

\paragraph{Multi-modal graph construction \& Score Propagation.}
For each sample in $\mathcal{D}_E$, we compute both visual and textual embeddings,
\vspace{-0.2cm}
\begin{equation}
h_i^{(v)} = f_v(x_i), \quad 
h_i^{(t)} = f_t(t_i),
\end{equation}
where $f_v$ is a visual encoder (e.g., CLIP ViT-L/14~\citep{radford2021learning}).
We then construct modality-specific graphs
$\mathcal{G}^{(v)}$ and $\mathcal{G}^{(t)}$
using $k$-NN with cosine similarity following a similar fashion in Eq.~\ref{eq:knn_graph}. This results in two graphs featuring both visual $\mathcal{G}^{(v)}$ and textual $\mathcal{G}^{(t)}$ modalities.

Given a query $q$, we initialize modality-specific relevance vectors
$I_0^{(m)}(q)$ by assigning value $1$ to the $k$ nearest neighbors of $q$ in modality $m\in\{v,t\}$ and $0$ to all other nodes.
Relevance is then propagated following the closed-form solution:
\vspace{-0.2cm}

\begin{equation}
    I^{(m)}_{(\infty)} = (1-\alpha)(\mathbf{I}-\alpha \hat{W}^{(m)})^{-1}I^{(m)}_{(0)}
\end{equation}

\noindent\textbf{Demonstration Selection and Inference}.
After convergence, we apply late fusion to combine the modality-specific relevance scores via,
\begin{equation}
I = \beta I^{(v)}_{(\infty)} + (1-\beta) I^{(t)}_{(\infty)},
\end{equation}
where $\beta \in [0,1]$ balances the visual and textual modalities.
For each query $q$, we rank all candidates in $\mathcal{D}_E$ by the fused relevance score $I$ and select the top-$k$ demonstrations:
\begin{equation}
DS(q) = \mathrm{Top}\text{-}k \left( \{ I_i \}_{x_i \in \mathcal{D}_E} \right).
\end{equation}
The selected demonstrations are then provided to the MLLM for inference:
\begin{equation}
\hat{a}_q = \mathcal{M}(x_q, q_q; DS(q)).
\end{equation}

\noindent\textbf{Efficient Incremental Inference.}
The static graph construction described above requires attaching  query samples to the graph. Without seeing all query samples, a naive way to incrementally do inference for a new query sample may need to rebuild the graph from scratch. However, for a graph with $N = |\mathcal{D}_E| + 1$ nodes, this incurs additional computational cost. To enable efficient  inference while still exploiting the manifold structure, we introduce a simple graph update mechanism.
Given an existing modality-specific graph
$\mathcal{G}^{(m)} = (\mathcal{V}^{(m)}, \mathcal{E}^{(m)})$
with adjacency matrix $W^{(m)} \in \mathbb{R}^{N \times N}$,
we insert a new query sample $x_q$ by measuring its similarity to all existing nodes:
\begin{equation}
s_{qi}^{(m)} = \langle h_q^{(m)}, h_i^{(m)} \rangle,
\quad \forall i \in \{1,\dots,|\mathcal{D}_E|\}.
\end{equation}

Rather than recomputing $k$-nearest neighbors for all nodes, we only remove the previous query sample from the graph and create edges for the top K nodes and connect them to the new query sample. This operation only requires calculating the inner product between query and $|\mathcal{D}_E|$ samples when each new query arrives.



\section{Experiments}

\vspace{-0.2cm}
\subsection{Experimental Setup}
\vspace{-0.2cm}

\noindent\textbf{Datasets}.
We evaluate on eight benchmarks spanning visual emotion recognition, scene text understanding, visual reasoning, and visual question answering. These tasks require diverse visual-textual understanding capabilities, from recognizing subtle emotional cues to complex spatial reasoning. \textbf{ Visual Emotion Recognition:} EmoSet~\citep{yang2023emoset} contains images annotated with eight emotion categories for emotion classification, while Emotion6~\citep{panda2018contemplating} provides six basic emotion labels collected via controlled user studies. \textbf{Scene Text Understanding:} TextOCR~\citep{zong2024vlicl} evaluates text detection and recognition in natural scene images. \textbf{Visual Reasoning:} MMStar~\citep{chen2024we} contains 1,500 challenge samples testing multi-modal reasoning with reduced language bias. MatchingMI~\citep{zong2024vlicl} requires cross-image matching and correspondence. CLEVR~\citep{zong2024vlicl} is a synthetic diagnostic benchmark for compositional reasoning including counting, comparison, and logical operations. \textbf{Visual Question Answering:} GQA~\citep{hudson2019gqa} features compositional questions grounded in scene graphs. OK-VQA~\citep{okvqa} evaluates open-domain VQA requiring external commonsense knowledge. For all datasets, we report accuracy as an evaluation metric.

\noindent\textbf{Competing Methods}.
We compare against seven baselines spanning both zero-shot approaches and few-shot retrieval methods. 
\textbf{Few-shot methods:}  \textit{Few-shot}~\cite{brown2020language} randomly samples labeled demonstrations. \textit{VICL}~\citep{zhou2024visual} uses retrieval and reranking with language-based demonstrations. \textit{Top-K+MDL}~\citep{wu2023self} ranks candidates by content similarity and minimum description length. \textit{MMICES}~\citep{chen2025can} performs two-stage selection via visual concept consistency and textual similarity. 
\textit{CVR-LLM}~\citep{li2024enhancing} generates context-aware visual descriptions through iterative refinement and conducts demonstration selection using multi-modal features. \textit{MAPLE}~\citep{chen2025maple} performs pseudo-labeling on unlabeled data and selects demonstrations from a graph constructed using textual questions only.
\textbf{Zero-shot methods:} \textit{Zero-shot} uses task instructions without demonstrations. \textit{Cola-Zero}~\citep{chen2023large} coordinates multiple VLMs via an LLM. 

\subsection{Implementation Details}
\vspace{-0.2cm}
\noindent\textbf{ICL Configurations.}
For each dataset, we randomly sample 15 labeled examples, 585 unlabeled examples as the candidate pool, and 300 test samples. From the unlabeled pool, we pseudo-label the top 45 high-relevance samples identified in Stage~1, forming an expanded candidate pool of 60 samples ($|\mathcal{D}_E| = 15 + 45$). For few-shot methods, both MAPLE and MAG select 10 demonstrations from 15 labeled + 585 unlabeled examples while all others select 10 demonstrations from 15 labeled ones.

\noindent\textbf{Text Description Generation.}
For all samples, we generate image descriptions using Gemini-2.0-Flash~\citep{comanici2025gemini25pushingfrontier} with the following prompt:
\textit{``Generate a description for the image, taking the question as guidance, elucidating both the visual content and the underlying purpose or intention depicted. Craft a clear and concise description that integrates details from the image, highlighting visual cues and semantic meaning which may be useful to get the answer.''}
These descriptions are used to construct textual representations in both Stage~1 and Stage~2.

\noindent\textbf{Models and hyperparameters.}
We adopt Gemini-2.0-Flash as the base MLLM. Contriever~\citep{izacard2022unsuperviseddenseinformationretrieval} is used for text encoding ($f_t$), and CLIP ViT-L/14~\citep{radford2021learning} for visual encoding ($f_v$). To investigate the generalization of method, we also evaluate against alternative MLLMs, including Gemini-3.0-Flash, GPT-4o, GPT-4.1, GPT-5 and Qwen models in the appendix. For graph construction and propagation, we set the number of nearest neighbors to $k_{nn}=5$, the propagation strength to $\alpha=0.9$, and the modality fusion weight to $\beta=0.5$. These hyperparameters are fixed across all experiments for consistency.

\subsection{Quantitave Evaluations}
\label{subsec:main_results}
\vspace{-0.2cm}

Table~\ref{tab:main_results} compares MAG against baselines under label-scarce conditions, reported with average and standard deviation over 3 random runs. We make the following observations from the results.

\begin{table*}[h!t!b]
\centering
\caption{Main results on multi-modal benchmarks under label-scarce settings with Gemini-2.0-Flash (mean accuracy $\pm$ standard dev \% across 3 runs).}
\label{tab:main_results}
\setlength{\tabcolsep}{2.5pt}
  \renewcommand{\arraystretch}{1.0}
  \resizebox{0.99\textwidth}{!}{
\begin{small}
\begin{sc}
\begin{tabular}{lcccccccc}
\toprule
Method & EmoSet & Emotion6 & TextOCR & MMStar & CLEVR & MatchingMI & GQA & OKVQA \\
\midrule
\multicolumn{9}{l}{\textit{Zero-shot Methods}} \\
Zero-shot & 60.1$\pm$4.2& 60.6$\pm$0.6& 56.1$\pm$1.2& 23.9$\pm$2.9& 59.8$\pm$2.5& 73.2$\pm$3.0& 57.6$\pm$0.5& 55.2$\pm$1.1\\
Cola-Zero~\cite{chen2023large}& 47.7$\pm$2.3& 49.4$\pm$0.5& 25.0$\pm$1.0& 34.2$\pm$2.5& 24.9$\pm$1.6& 18.2$\pm$3.3& 51.4$\pm$2.8& 49.6$\pm$1.3\\
\midrule
\multicolumn{9}{l}{\textit{Few-shot Methods}} \\
Few-shot~\cite{brown2020language} & 61.0$\pm$2.9& 63.0$\pm$1.9& 67.6$\pm$3.9& 65.8$\pm$4.8& 61.3$\pm$3.9& 96.7$\pm$0.7& 54.2$\pm$0.7& 54.8$\pm$2.3\\
VICL~\cite{zhou2024visual} & 62.6$\pm$1.6& 61.5$\pm$4.5& 67.0$\pm$3.7& 64.1$\pm$3.1& 50.6$\pm$1.5& 92.0$\pm$3.7& 55.1$\pm$1.0& 56.3$\pm$1.5\\
Top-K+MDL~\cite{wu2023self} & 59.6$\pm$1.3& 62.8$\pm$1.7& 64.3$\pm$0.9& 62.3$\pm$2.0& 57.8$\pm$2.4& 94.1$\pm$0.8& 56.2$\pm$1.1& 54.7$\pm$1.4\\
MMICES~\cite{chen2025can}) & 60.4$\pm$3.7& 62.9$\pm$0.9& 66.3$\pm$1.4& 60.8$\pm$2.6& 61.6$\pm$2.2& 94.2$\pm$1.5& 53.1$\pm$1.3& 54.9$\pm$1.8\\
CVR-LLM~\cite{li2024enhancing}& 74.8$\pm$1.5& 62.5$\pm$2.3& 87.9$\pm$1.2& 64.4$\pm$2.5& 90.2$\pm$1.8& 97.6$\pm$0.9& 52.3$\pm$1.4& 46.7$\pm$2.0\\
MAPLE~\cite{chen2025maple} & 72.4$\pm$1.2& 61.3$\pm$1.3& 84.0$\pm$0.8& 64.8$\pm$1.7& 89.3$\pm$0.6& 96.2$\pm$0.4& 55.7$\pm$2.1& 54.3$\pm$2.3\\
\textbf{MAG}~(Ours)& \textbf{76.3$\pm$1.4}& \textbf{64.3$\pm$1.4}& \textbf{91.4$\pm$0.6}& \textbf{67.9$\pm$2.1}& \textbf{92.1$\pm$0.5}& \textbf{99.8$\pm$0.2}& \textbf{59.0$\pm$3.9}& \textbf{57.6$\pm$2.1}\\
\bottomrule
\end{tabular}
\end{sc}
\end{small}
}
\end{table*}

\noindent\textbf{Unlabeled data provides substantial gains under label scarcity.} MAG consistently outperforms all baselines across eight benchmarks using only 15 labeled samples. Compared with the strongest retrieval baseline (MMICES), MAG achieves notable improvements on challenging tasks, including +30.5\% on CLEVR, +7.1\% on MMStar, and +25.1\% on TextOCR. Relative to standard few-shot prompting, MAG further improves TextOCR (+23.8\%), CLEVR (+30.8\%), and EmoSet (+15.3\%). These results demonstrate the effectiveness of leveraging unlabeled data through relevance-guided pseudo-labeling and graph propagation.

\noindent\textbf{Graph-based propagation outperforms similarity-based retrieval.} Compared with similarity-based retrieval methods such as VICL and Top-K+MDL, MAG consistently achieves stronger performance by modeling global relational structure among samples. For instance, MAG surpasses Top-K+MDL by +34.3\% on CLEVR and +27.1\% on TextOCR, validating that graph connectivity provides more reliable supervision than local similarity alone under limited labeled data.

\noindent\textbf{Our method complements description-based approaches.} Although CVR-LLM achieves competitive performance through iterative description refinement, MAG still improves results on most benchmarks, including +3.5\% on EmoSet, +3.5\% on TextOCR, +6.7\% on GQA, and +10.9\% on OKVQA. Unlike CVR-LLM, which shows unstable cross-task generalization, MAG maintains consistently strong performance, suggesting that pseudo-label expansion and graph propagation provide a more robust way to exploit unlabeled data.

\noindent\textbf{Stronger performance than single modality manifold method.} Compared with the single modality manifold based method (MAPLE), we observe consistent improvement across all datasets. Despite both methods selecting demonstrations from the same labeled + unlabeled examples, our method excels by exploiting multiple modalities and principled selection strategy.

Overall, these results confirm that MAG enables effective multi-modal ICL by incorporating unlabeled data, achieving strong and stable performance. Qualitative examples demonstrating improved demonstration selection are provided in Appendix~\ref{appendix:qualitative}.

\subsection{Ablation Study}
\vspace{-0.2cm}

We conduct ablation studies to validate the key components of MAG. Table~\ref{tab:ablation} evaluates three design choices: pseudo-labeling strategy, demonstration selection method, and modality usage. We report mean and standard deviation across 3 runs and draw the following observations.

\textbf{Manifold-guided pseudo-labeling is critical, and naive pseudo-labeling can be harmful.}
Comparing three strategies, we observe a clear trend: (1) removing pseudo-labeling yields 72.0\% on EmoSet, (2) adding randomly selected pseudo-labels provides only marginal gains on EmoSet (73.1\%) and even degrades performance on some tasks (e.g., GQA: 54.7\% vs. 56.4\%), while (3) our manifold-guided pseudo-labeling significantly improves performance across all benchmarks (76.3\% on EmoSet, 91.4\% on TextOCR, 67.9\% on MMStar, 58.0\% on GQA). 
This indicates that pseudo-label quality, rather than quantity, is the determining factor. Naively introducing pseudo-labels can inject noise and offset potential gains, whereas our relevance-guided selection effectively identifies high-value samples, leading to consistent improvements.

\textbf{Graph-based demonstration selection consistently outperforms heuristic strategies.}
Compared to random selection, graph-based selection yields substantial gains (+3.5\% on EmoSet, +2.5\% on TextOCR, +1.8\% on GQA). Replacing graph propagation with TopK similarity leads to smaller improvements and consistently underperforms the full method.
This suggests that selecting demonstrations purely based on query-example similarity is insufficient. Instead, graph propagation captures higher-order relationships among samples, enabling retrieval of globally relevant and representative demonstrations. This advantage becomes more pronounced in low-resource settings where labeled examples are sparse and potentially biased.

\textbf{Both modalities are necessary and complementary.}
Removing either modality results in significant performance degradation. Text-only graphs (–img) particularly hurt visually grounded reasoning (GQA: 44.2\% vs. 58.0\%), while visual-only graphs (–desc) fail on semantically intensive tasks (EmoSet: 60.2\% vs. 76.3\%, TextOCR: 67.0\% vs. 91.4\%).
Notably, visual-only selection introduces misleading context due to superficial visual similarity, leading to large drops on language-heavy tasks. These results highlight that visual and textual modalities encode complementary structures, and effective demonstration selection requires jointly modeling both.

\begin{table*}[h]
\centering
\vspace{-0.5cm}
\caption{Ablation study. We evaluate the contribution of (1) relevance-guided pseudo-labeling (MAPLE vs. Random vs. None), (2) graph-based propagation (Graph vs. TopK vs. Random), and (3) multi-modal representations (with/without images and descriptions).}
\label{tab:ablation}
\setlength{\tabcolsep}{2.5pt}
  \renewcommand{\arraystretch}{1.0}
  \resizebox{0.90\textwidth}{!}{
\begin{small}
\begin{sc}
\begin{tabular}{@{}>{\raggedright\arraybackslash}p{3cm}cccccc@{}}
\toprule
Method Variant & Pseudo-label & Demo. Selection & EmoSet & TextOCR & MMStar & GQA \\
\midrule
\multicolumn{7}{l}{\textit{Ablating pseudo-labeling strategy}} \\
Ours (–pseudo)    & -  & MMGraph & 72.0$\pm$0.3 & 88.1$\pm$2.7 & 66.5$\pm$0.7 & 56.4$\pm$1.4 \\
Ours (+rand)      & Random  & MMGraph & 73.1$\pm$2.7 & 88.0$\pm$3.6 & 66.4$\pm$1.6 & 54.7$\pm$2.1 \\
Ours (full)       & Manifold & MMGraph & \textbf{76.3$\pm$1.4} & \textbf{91.4$\pm$0.6} & \textbf{67.9$\pm$2.1} & \textbf{58.0$\pm$3.9} \\
\midrule
\multicolumn{7}{l}{\textit{Ablating demonstration selection}} \\
Ours (Rand)       & Manifold & Random  & 72.8$\pm$2.0 & 88.9$\pm$1.9 & 65.1$\pm$4.3 & 56.2$\pm$1.9 \\
Ours (TopK)       & Manifold & TopK  & 74.1$\pm$1.9 & 89.0$\pm$1.0 & 67.6$\pm$3.1 & 56.1$\pm$1.9 \\
Ours (full)       & Manifold & MMGraph & \textbf{76.3$\pm$1.4} & \textbf{91.4$\pm$0.6} & \textbf{67.9$\pm$2.1} & \textbf{58.0$\pm$3.9} \\
\midrule
\multicolumn{7}{l}{\textit{Ablating modalities}} \\
Ours (–img)       & Manifold & Graph (text only) & 75.2$\pm$1.0 & 90.4$\pm$0.5 & 67.4$\pm$2.5 & 44.2$\pm$2.6 \\
Ours (–desc)      & Manifold & Graph (visual only) & 60.2$\pm$3.6 & 67.0$\pm$2.4 & 65.9$\pm$2.9 & 52.4$\pm$1.4 \\
Ours (full)       & Manifold & MMGraph (both) & \textbf{76.3$\pm$1.4} & \textbf{91.4$\pm$0.6} & \textbf{67.9$\pm$2.1} & \textbf{58.0$\pm$3.9} \\
\bottomrule
\end{tabular}
\end{sc}
\end{small}
}
\vspace{-0.3cm}
\end{table*}

\subsection{Modality Analysis for Stage 1}
\vspace{-0.2cm}

We further evaluate MAG using different modality representations for Stage 1 pseudo-label selection. Table~\ref{tab:modality_s1} shows results for textual-only, visual-only, and combined representations.

\textbf{Textual representations provide robust cross-task performance.} Textual-only achieves best performance on 6 of 8 datasets, particularly excelling on emotion recognition (EmoSet: 77.0\%, Emotion6: 65.0\%), scene text understanding (TextOCR: 91.7\%), and visual reasoning (MMStar: 70.3\%, GQA: 62.3\%). This suggests that MLLM-generated image descriptions capture semantic concepts that are well-aligned with these task requirements.

\textbf{Visual representations excel slightly on spatial reasoning.} Visual-only achieves marginally higher accuracy on spatially-intensive tasks (CLEVR: 92.3\% vs. 92.0\%, OKVQA: 58.0\% vs. 56.7\%), where geometric relationships are critical. However, the performance gap is small (+0.3\% to +1.3\%).

\textbf{Choose textual representation only for Stage 1.} Given that textual representations provide strong and stable performance across diverse task types, we use textual-only for Stage 1 in all experiments. Importantly, even on datasets where visual representations provide small gains in Stage 1 isolation, our full method with textual Stage 1 still substantially outperforms all baselines (e.g., 92.0\% vs. 90.3\% baseline on CLEVR, 56.7\% vs. 54.0\% baseline on OKVQA in Table~\ref{tab:main_results}). This demonstrates that Stage 2's multi-modal propagation effectively compensates for Stage 1 modality choice, as it incorporates both visual and textual information through complementary graphs.

\begin{table*}[h!t!b]
\centering
\vspace{-0.4cm}
\caption{Performance using different Stage 1 modalities for pseudo-label selection (accuracy \%). Textual representations provide robust performance across tasks. Visual representations achieve slightly higher accuracy on spatially-intensive tasks (CLEVR, OKVQA) but the gap is small (+0.3\% to +1.3\%). We use textual representations as default for consistency.}
\label{tab:modality_s1}
\setlength{\tabcolsep}{2.5pt}
  \renewcommand{\arraystretch}{1.0}
  \resizebox{0.94\textwidth}{!}{
\begin{small}
\begin{sc}
\begin{tabular}{lcccccccc}
\toprule
Stage 1 Modality & EmoSet & Emotion6 & TextOCR & MMStar & CLEVR & MatchingMI & GQA & OKVQA \\
\midrule
Textual only & \textbf{77.0} & \textbf{65.0} & \textbf{91.7} & \textbf{70.3} & 92.0 & 99.7 & \textbf{62.3} & 56.7 \\
Visual only & 75.0 & 60.7 & 85.3 & 62.3 & \textbf{92.3} & 99.7 & 57.3 & \textbf{58.0} \\
Textual \& Visual & 72.7 & 59.7 & 87.0 & 61.7 & 91.3 & 99.7 & 57.7 & 54.7 \\
\bottomrule
\end{tabular}
\end{sc}
\end{small}
}
\vspace{-0.4cm}
\end{table*}

\subsection{Multi-Modal Fusion Strategy}
\vspace{-0.2cm}

We compare three modality fusion strategies at Stage 2 for computing relevance score: early fusion (concatenate the textual embeddings and visual embeddings before graph construction), mid fusion (utilize the weighted sum of adjacency matrix for relevance propagation), and late fusion (do relevance propagation of two modalities separately, then average these two relevance scores). Table~\ref{tab:fusion} shows that late fusion consistently achieves the best performance across all benchmarks, with particularly large gains on reasoning-intensive tasks (MMStar: +6.6\% over early fusion, GQA: +2.3\%). Early fusion performs worst, suggesting that directly mixing heterogeneous visual and textual features before relational modeling weakens modality-specific structure learning. Late fusion preserves complementary relational signals by allowing each modality's graph to capture its own semantic structure before combining scores, which is particularly beneficial for tasks requiring cross-modal reasoning.

\begin{table*}[h!t!b]
\centering
\vspace{-0.2cm}
\caption{Performance using different fusion strategies for combining visual and textual relevance scores (accuracy \%). Late fusion consistently outperforms early and mid fusion by preserving modality-specific graph structures during propagation.}
\label{tab:fusion}
\setlength{\tabcolsep}{2.5pt}
  \renewcommand{\arraystretch}{1.0}
  \resizebox{0.94\textwidth}{!}{
\begin{small}
\begin{sc}
\begin{tabular}{lcccccccc}
\toprule
Fusion Strategy & EmoSet & Emotion6 & TextOCR & MMStar & CLEVR & MatchingMI & GQA & OKVQA \\
\midrule
Early fusion & 76.0 & 60.0 & 86.3 & 63.7 & 89.3 & 99.3 & 60.0 & 53.7 \\
Mid fusion & \textbf{77.0} & 60.7 & 88.0 & 64.3 & 91.0 & 99.3 & 56.7 & 52.0 \\
Late fusion & \textbf{77.0} & \textbf{65.0} & \textbf{91.7} & \textbf{70.3} & \textbf{92.0} & \textbf{99.7} & \textbf{62.3} & \textbf{56.7} \\
\bottomrule
\end{tabular}
\end{sc}
\end{small}
}
\vspace{-0.4cm}
\end{table*}

\subsection{Hyperparameter Sensitivity}

\begin{figure*}[!t]
\vspace{-0.0cm}\begin{center}
\includegraphics[width=1.02\textwidth]{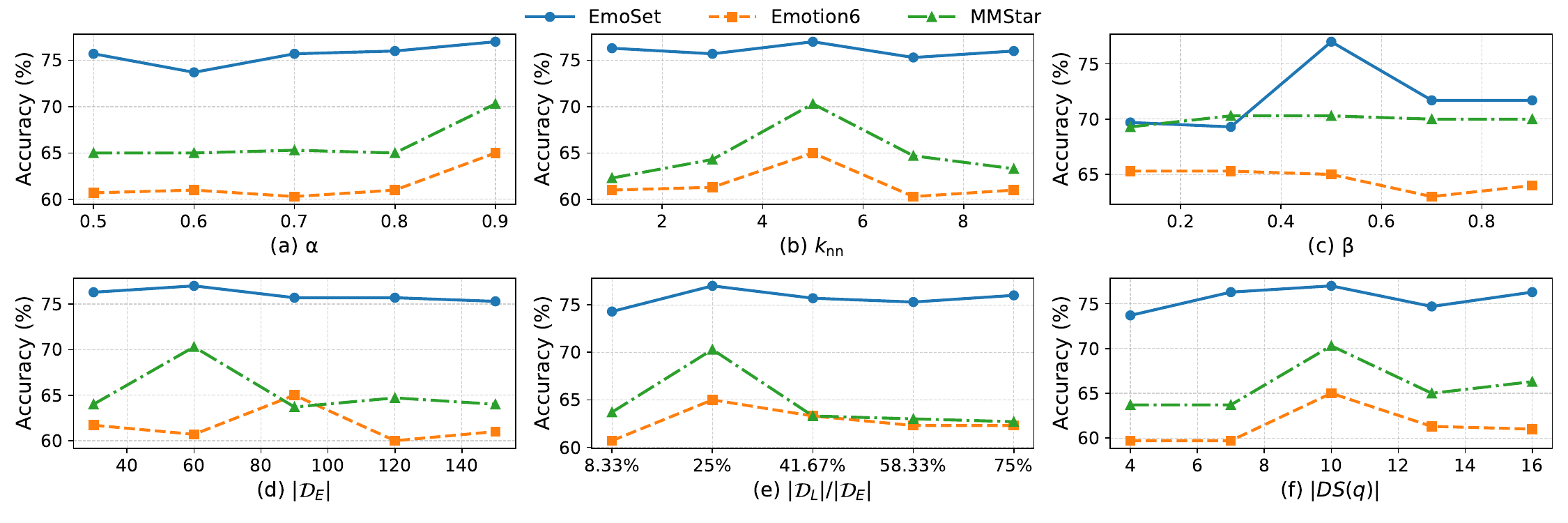}
\vspace{-0.8cm}
\caption{Sensitivity analysis across three datasets (EmoSet, Emotion6, MMStar). Performance is stable across wide ranges of $\alpha$, $\beta$, $k_{nn}$, pool size, and label rate, with clear optimal configurations: $\alpha \approx 0.9$, $k_{nn}=5$, $\beta=0.5$, pool size of 60, label rate of 25\%, and 10 demonstrations.}
\label{fig:hyperparam}
\end{center}
\vspace{-0.7cm}
\end{figure*}

We evaluate MAG's sensitivity to key hyperparameters. Figure~\ref{fig:hyperparam} shows results across EmoSet, Emotion6, and MMStar for (a) propagation strength~($\alpha$), (b) modality fusion weight~($\beta$), (c) nearest neighbors ($k_{nn}$), (d) candidate pool size~($|\mathcal{D}_E|$), (e) label rate within candidate pool size~($|\mathcal{D}_L|/|\mathcal{D}_E|$), and (f) number of demonstrations~($|DS(q)|$). Performance remains stable across wide ranges for most parameters, with clear optimal configurations emerging: $\alpha \approx 0.9$ (emphasizing propagation over initialization), $k_{nn}=5$ (moderate neighborhood size), $\beta=0.5$ (balanced modality fusion), pool size of 60 (15 labeled + 45 pseudo-labeled), label rate of 25\% (demonstrating label efficiency), and 10 demonstrations (beyond which performance plateaus or degrades). Notably, the label rate analysis shows that performance peaks at only 25\% labeled data, validating that our relevance-guided pseudo-labeling effectively leverages unlabeled samples. The framework demonstrates strong robustness to hyperparameter choices while maintaining consistent optimal configurations across diverse tasks.

\vspace{-0.2cm}

\section{Conclusion}
\vspace{-0.2cm}
We introduced MAG, a semi-supervised framework for multi-modal in-context learning that effectively leverages unlabeled data under label-scarce settings. By modeling demonstration selection as relevance propagation on multi-modal graphs, MAG identifies a compact set of high-impact unlabeled samples for pseudo-labeling and selects query-conditioned demonstrations that are globally relevant across visual and textual modalities. This design enables efficient use of unlabeled data while respecting inference-time and context-length constraints.
Extensive experiments across eight diverse multi-modal benchmarks show that MAG consistently outperforms strong few-shot and semi-supervised baselines, achieving substantial gains with a limited pseudo-labeling budget. Our analysis highlights the complementary roles of textual and visual manifolds in ICL and demonstrates that principled graph-based selection is key to scalable multi-modal prompting. We hope MAG provides a practical foundation for exploiting unlabeled data in future multi-modal ICL systems.

\bibliography{our_references}

@String(CVPR  = {IEEE/CVF Conference on Computer Vision and Pattern Recognition})

@String(ICCV  = {IEEE/CVF International Conference on Computer Vision})

@String(ECCV  = {European Conference on Computer Vision})

@String(WACV  = {IEEE/CVF Winter Conference on Applications of Computer Vision})

@String(NIPS  = {Advances in Neural Information Processing Systems})

@String(ICLR  = {International Conference on Learning Representations})

@String(ICML  = {International Conference on Machine Learning})

@String(AAAI  = {AAAI Conference on Artificial Intelligence})

@String(ACL    = {Annual Meeting of the Association for Computational Linguistics})

@String(EMNLP  = {Conference on Empirical Methods in Natural Language Processing})

@String(NAACL = {Annual Conference of the North American Chapter of the Association for Computational Linguistics})

@inproceedings{an2023context,
  title={How do in-context examples affect compositional generalization?},
  author={An, Shengnan and Lin, Zeqi and Fu, Qiang and Chen, Bei and Zheng, Nanning and Lou, Jian-Guang and Zhang, Dongmei},
  booktitle=ACL,
  year={2023}
}

@inproceedings{wu2023self,
  title={Self-adaptive in-context learning: An information compression perspective for in-context example selection and ordering},
  author={Wu, Zhiyong and Wang, Yaoxiang and Ye, Jiacheng and Kong, Lingpeng},
  booktitle=ACL,
  year={2023}
}

@inproceedings{levy2023diverse,
  title={Diverse demonstrations improve in-context compositional generalization},
  author={Levy, Itay and Bogin, Ben and Berant, Jonathan},
  booktitle=ACL,
  year={2023}
}

@inproceedings{chen2025maple,
  title={MAPLE: Many-Shot Adaptive Pseudo-Labeling for In-Context Learning},
  author={Chen, Zihan and Wang, Song and Tan, Zhen and Li, Jundong and Shen, Cong},
  booktitle=ICML,
  year={2025}
}

@inproceedings{liu2022few,
  title={Few-shot parameter-efficient fine-tuning is better and cheaper than in-context learning},
  author={Liu, Haokun and Tam, Derek and Muqeeth, Mohammed and Mohta, Jay and Huang, Tenghao and Bansal, Mohit and Raffel, Colin A},
  booktitle=NIPS,
  year={2022}
}

@inproceedings{ye2023compositional,
  title={Compositional exemplars for in-context learning},
  author={Ye, Jiacheng and Wu, Zhiyong and Feng, Jiangtao and Yu, Tao and Kong, Lingpeng},
  booktitle=ICML,
  year={2023}
}

@inproceedings{li2024configure,
  title={How to configure good in-context sequence for visual question answering},
  author={Li, Li and Peng, Jiawei and Chen, Huiyi and Gao, Chongyang and Yang, Xu},
  booktitle=CVPR,
  year={2024}
}

@inproceedings{chen2025can,
  title={Can Multimodal Large Language Models Truly Perform Multimodal In-Context Learning?},
  author={Chen, Shuo and Han, Zhen and He, Bailan and Liu, Jianzhe and Buckley, Mark and Qin, Yao and Torr, Philip and Tresp, Volker and Gu, Jindong},
  booktitle=WACV,
  year={2025}
}

@article{zhou2024visual,
  title={Visual in-context learning for large vision-language models},
  author={Zhou, Yucheng and Li, Xiang and Wang, Qianning and Shen, Jianbing},
  journal={arXiv preprint arXiv:2402.11574},
  year={2024}
}

@inproceedings{li2024context,
  title={In-context compositional generalization for large vision-language models},
  author={Li, Chuanhao and Jing, Chenchen and Li, Zhen and Zhai, Mingliang and Wu, Yuwei and Jia, Yunde},
  booktitle=EMNLP,
  year={2024}
}

@inproceedings{chen2023large,
  title={Large language models are visual reasoning coordinators},
  author={Chen, Liangyu and Li, Bo and Shen, Sheng and Yang, Jingkang and Li, Chunyuan and Keutzer, Kurt and Darrell, Trevor and Liu, Ziwei},
  booktitle=NIPS,
  year={2023}
}

@article{rahman2025mpcar,
  title={MPCAR: Multi-Perspective Contextual Augmentation for Enhanced Visual Reasoning in Large Vision-Language Models},
  author={Rahman, Amirul and Xu, Qiang and Huang, Xueying},
  journal={arXiv preprint arXiv:2508.12400},
  year={2025}
}

@article{deng2024enhancing,
  title={Enhancing large vision language models with self-training on image comprehension},
  author={Deng, Yihe and Lu, Pan and Yin, Fan and Hu, Ziniu and Shen, Sheng and Gu, Quanquan and Zou, James Y and Chang, Kai-Wei and Wang, Wei},
  journal=NIPS,
  year={2024}
}

@inproceedings{li2025taco,
  title={Taco: Enhancing multimodal in-context learning via task mapping-guided sequence configuration},
  author={Li, Yanshu and Yang, Jianjiang and Yun, Tian and Feng, Pinyuan and Huang, Jinfa and Tang, Ruixiang},
  booktitle=EMNLP,
  year={2025}
}

@inproceedings{li2024enhancing,
  title={Enhancing advanced visual reasoning ability of large language models},
  author={Li, Zhiyuan and Liu, Dongnan and Zhang, Chaoyi and Wang, Heng and Xue, Tengfei and Cai, Weidong},
  booktitle=EMNLP,
  year={2024}
}

@inproceedings{iscen2017efficient,
  title={Efficient diffusion on region manifolds: Recovering small objects with compact cnn representations},
  author={Iscen, Ahmet and Tolias, Giorgos and Avrithis, Yannis and Furon, Teddy and Chum, Ondrej},
  booktitle={IEEE conference on computer vision and pattern recognition},
  year={2017}
}

@inproceedings{zhang2020hypergraph,
  title={Hypergraph label propagation network},
  author={Zhang, Yubo and Wang, Nan and Chen, Yufeng and Zou, Changqing and Wan, Hai and Zhao, Xinbin and Gao, Yue},
  booktitle=AAAI,
  year={2020}
}

@inproceedings{chen2025cross,
  title={From Cross-Task Examples to In-Task Prompts: A Graph-Based Pseudo-Labeling Framework for In-context Learning},
  author={Chen, Zihan and Wang, Song and Fu, Xingbo and Shi, Chengshuai and Lei, Zhenyu and Shen, Cong and Li, Jundong},
  booktitle=EMNLP,
  year={2025}
}

@inproceedings{brown2020language,
  title={Language models are few-shot learners},
  author={Brown, Tom and Mann, Benjamin and Ryder, Nick and others},
  booktitle=NIPS,
  year={2020}
}

@article{peebles2022learning,
  title={Learning to learn with generative models of neural network checkpoints},
  author={Peebles, William and Radosavovic, Ilija and Brooks, Tim and Efros, Alexei A and Malik, Jitendra},
  journal={arXiv preprint arXiv:2209.12892},
  year={2022}
}

@inproceedings{garg2022can,
  title={What can transformers learn in-context? a case study of simple function classes},
  author={Garg, Shivam and Tsipras, Dimitris and Liang, Percy S and Valiant, Gregory},
  booktitle=NIPS,
  year={2022}
}

@article{xie2022explanation,
  title={An explanation of in-context learning as implicit bayesian inference},
  author={Xie, Sang Michael and Raghunathan, Aditi and Liang, Percy and Ma, Tengyu},
  journal={arXiv preprint arXiv:2111.02080},
  year={2021}
}

@inproceedings{lu2022fantastically,
  title={Fantastically Ordered Prompts and Where to Find Them: Overcoming Few-Shot Prompt Order Sensitivity},
  author={Lu, Yao and Bartolo, Max and Moore, Alastair and Riedel, Sebastian and Stenetorp, Pontus},
  booktitle=ACL,
  year={2022}
}

@inproceedings{rubin2022learning,
  title={Learning To Retrieve Prompts for In-context Learning},
  author={Rubin, Ohad and Herzig, Jonathan and Berant, Jonathan},
  booktitle=NAACL,
  year={2022}
}

@inproceedings{zhou2022large,
  title={Large language models are human-level prompt engineers},
  author={Zhou, Yongchao and Muresanu, Andrei Ioan and Han, Ziwen and Paster, Keiran and Pitis, Silviu and Chan, Harris and Ba, Jimmy},
  booktitle=ICLR,
  year={2022}
}

@inproceedings{pryzant2023automatic,
  title={Automatic prompt optimization with" gradient descent" and beam search},
  author={Pryzant, Reid and Iter, Dan and Li, Jerry and Lee, Yin Tat and Zhu, Chenguang and Zeng, Michael},
  booktitle=ACL,
  year={2023}
}

@inproceedings{radford2021learning,
  title={Learning Transferable Visual Models From Natural Language Supervision},
  author={Radford, Alec and Kim, Jong Wook and Hallacy, Chris and Ramesh, Aditya and Goh, Gabriel and Agarwal, Sandhini and Sastry, Girish and Askell, Amanda and Mishkin, Pamela and Clark, Jack and others},
  booktitle=ICML,
  year={2021}
}

@inproceedings{jia2021scaling,
  title={Scaling Up Visual and Vision-Language Representation Learning With Noisy Text Supervision},
  author={Jia, Chao and Yang, Yinfei and Xia, Ye and Chen, Yi-Ting and Parekh, Zarana and Pham, Hieu and Le, Quoc and Sung, Yun-Hsuan and Li, Zhen and Duerig, Tom},
  booktitle=ICML,
  year={2021}
}

@inproceedings{li2022blip,
  title={BLIP: Bootstrapped Language-Image Pre-training for Unified Vision-Language Understanding and Generation},
  author={Li, Junnan and et al.},
  booktitle=ICML,
  year={2022}
}

@inproceedings{alayrac2022flamingo,
  title={Flamingo: a Visual Language Model for Few-Shot Learning},
  author={Alayrac, Jean-Baptiste and Donahue, Jeff and Luc, Pauline and Miech, Antoine and Barr, Iain and Hasson, Yana and Lenc, Karel and Mensch, Arthur and Millican, Katherine and Reynolds, Malcolm and others},
  booktitle=NIPS,
  year={2022}
}

@article{openai2023gpt4,
  title={GPT-4 Technical Report},
  author={OpenAI},
  journal={arXiv preprint arXiv:2303.08774},
  year={2023}
}

@inproceedings{liu2023llava,
  title={Visual Instruction Tuning},
  author={Liu, Haotian and Li, Chunyuan and Wu, Qingyang and Lee, Yong Jae},
  booktitle=NIPS,
  year={2023}
}

@inproceedings{zhu2002learning,
  title={Learning from Labeled and Unlabeled Data with Label Propagation},
  author={Zhu, Xiaojin and Ghahramani, Zoubin and Lafferty, John},
  booktitle=ICML,
  year={2002}
}

@inproceedings{zhou2004learning,
  title={Learning with Local and Global Consistency},
  author={Zhou, Dengyong and Bousquet, Olivier and Lal, Thomas and Weston, Jason and Sch{\"o}lkopf, Bernhard},
  booktitle=NIPS,
  year={2004}
}

@inproceedings{iscen2019label,
  title={Label Propagation for Deep Semi-supervised Learning},
  author={Iscen, Ahmet and Tolias, Giorgos and Avrithis, Yannis and Chum, Ondrej},
  booktitle=CVPR,
  year={2019}
}

@inproceedings{basak2023pseudo,
  title={Pseudo-label guided contrastive learning for semi-supervised medical image segmentation},
  author={Basak, Hritam and Yin, Zhaozheng},
  booktitle=CVPR,
  year={2023}
}

@inproceedings{zheng2023simmatchv2,
  title={Simmatchv2: Semi-supervised learning with graph consistency},
  author={Zheng, Mingkai and You, Shan and Huang, Lang and Luo, Chen and Wang, Fei and Qian, Chen and Xu, Chang},
  booktitle=ICCV,
  year={2023}
}

@inproceedings{chen2025provoking,
  title={Provoking Multi-modal Few-Shot LVLM via Exploration-Exploitation In-Context Learning},
  author={Chen, Cheng and Zhai, Yunpeng and Zhao, Yifan and Gao, Jinyang and Ding, Bolin and Li, Jia},
  booktitle=CVPR,
  year={2025}
}

@inproceedings{zhu2003semi,
  title={Semi-supervised learning using gaussian fields and harmonic functions},
  author={Zhu, Xiaojin and Ghahramani, Zoubin and Lafferty, John D},
  booktitle=ICML,
  year={2003}
}

@inproceedings{tarvainen2017mean,
  title={Mean teachers are better role models: Weight-averaged consistency targets improve semi-supervised deep learning results},
  author={Tarvainen, Antti and Valpola, Harri},
  booktitle=NIPS,
  year={2017}
}

@inproceedings{sohn2020fixmatch,
  title={Fixmatch: Simplifying semi-supervised learning with consistency and confidence},
  author={Sohn, Kihyuk and Berthelot, David and Carlini, Nicholas and Zhang, Zizhao and Zhang, Han and Raffel, Colin A and Cubuk, Ekin Dogus and Kurakin, Alexey and Li, Chun-Liang},
  booktitle=NIPS,
  year={2020}
}

@inproceedings{yang2023emoset,
    title={A Large-scale Visual Emotion Dataset with Rich Attributes},
    author={Yang, Jingyuan and Huang, Qirui and Ding, Tingting and Lischinski, Dani and Cohen-Or, Danny and Huang, Hui},
    booktitle=ICCV,
    year={2023}
}

@inproceedings{panda2018contemplating,
    title={Contemplating Visual Emotions: Understanding and Overcoming Dataset Bias},
    author={Panda, Rameswar and Zhang, Jianming and Li, Haoxiang and Lee, Joon-Young and Lu, Xin and Roy-Chowdhury, Amit K},
    booktitle = ECCV,
    year={2018}
}

@article{zong2024vlicl,
    title={VL-ICL Bench: The Devil in the Details of Multimodal In-Context Learning},
    author={Zong, Yongshuo and Bohdal, Ondrej and Hospedales, Timothy},
    journal={arXiv preprint arXiv:2403.13164},
    year={2024}
}

@inproceedings{chen2024we,
    title={Are We on the Right Way for Evaluating Large Vision-Language Models?},
    author={Chen, Lin and Li, Jinsong and Dong, Xiaoyi and Zhang, Pan and Zang, Yuhang and Chen, Zehui and Duan, Haodong and Wang, Jiaqi and Qiao, Yu and Lin, Dahua and others},
    booktitle=NIPS,
    year={2024}
}

@inproceedings{hudson2019gqa,
  title={Gqa: A new dataset for real-world visual reasoning and compositional question answering},
  author={Hudson, Drew A and Manning, Christopher D},
  booktitle=CVPR,
  year={2019}
}

@inproceedings{okvqa,
    author = {Kenneth Marino and Mohammad Rastegari and Ali Farhadi and Roozbeh Mottaghi},
    title = {OK-VQA: A Visual Question Answering Benchmark Requiring External Knowledge},
    booktitle = CVPR,
    year = {2019},
}

@article{comanici2025gemini25pushingfrontier,
     title={Gemini 2.5: Pushing the frontier with advanced reasoning, multimodality, long context, and next generation agentic capabilities},
  author={Comanici, Gheorghe and Bieber, Eric and Schaekermann, Mike and Pasupat, Ice and Sachdeva, Noveen and Dhillon, Inderjit and Blistein, Marcel and Ram, Ori and Zhang, Dan and Rosen, Evan and others},
  journal={arXiv preprint arXiv:2507.06261},
  year={2025}
}

@article{izacard2022unsuperviseddenseinformationretrieval,
      title={Unsupervised dense information retrieval with contrastive learning},
  author={Izacard, Gautier and Caron, Mathilde and Hosseini, Lucas and Riedel, Sebastian and Bojanowski, Piotr and Joulin, Armand and Grave, Edouard},
  journal={arXiv preprint arXiv:2112.09118},
  year={2021}
}
\bibliographystyle{plainnat}

\newpage
\appendix
\onecolumn
\section*{Appendix}
\section{Algorithm Psuedo Code}\label{app:algo}

Algorithm~\ref{alg:MAG} summarizes the overall procedure of MAG, including the generation of the construction of the expanded pool, the final query-conditioned selection and prediction process.
\begin{algorithm}[h]
\caption{MAG}
\label{alg:MAG}
\begin{algorithmic}[1]
\STATE \textbf{Input:} Labeled data $\mathcal{D}_L$, unlabeled data $\mathcal{D}_U$, query $x_q$, $q_q$
\STATE \textbf{Output:} Predicted answer $\hat{a}_q$
\STATE 
\STATE \textit{// Generate textual descriptions for all images}
\FOR{$x_i \in \mathcal{D}_L \cup \mathcal{D}_U$}
\STATE $d_i \gets \mathcal{M}_{\text{desc}}(x_i)$ \quad \textit{// Generate image description}
\ENDFOR
\STATE 
\STATE \textit{// Stage 1: Relevance-guided pseudo-labeling}
\STATE Build textual graph $\mathcal{G}^{(t)}$ over $\mathcal{D}_L \cup \mathcal{D}_U$ using text embeddings
\STATE Compute relevance scores $I_j$ for $x_j \in \mathcal{D}_U$ via Eq. (4)
\FOR{$x_j \in \text{Top-}K\left(\{I_j\}_{x_j \in \mathcal{D}_U}\right)$}
\STATE Pseudo-label: $\hat{a}_j \gets \mathcal{M}(x_j, q_j; \mathcal{D}_L)$
\STATE Add $(x_j, \hat{a}_j)$ to $\mathcal{D}_P$
\ENDFOR
\STATE Expand pool: $\mathcal{D}_E \gets \mathcal{D}_L \cup \mathcal{D}_P$
\STATE 
\STATE \textit{// Stage 2: Multi-modal query-conditioned selection}
\STATE Build visual graph $\mathcal{G}^{(v)}$ (image embeddings) and textual graph $\mathcal{G}^{(t)}$ (text embeddings) over $\mathcal{D}_E$
\STATE Propagate score $I_{(\infty)}^{(m)}$ via Eq~(7)
\STATE Late Fusion: $I \gets \beta I^{(v)} + (1-\beta) I^{(t)}$
\STATE Select: $DS(q) \gets \text{Top-}k\left(\{I_i\}_{i=1}^{|\mathcal{D}_E|}\right)$
\STATE Predict: $\hat{a}_q \gets \mathcal{M}(x_q,q_q; DS(q))$
\STATE \textbf{return} $\hat{a}_q$
\end{algorithmic}
\end{algorithm}

\clearpage
\newpage

\section{Additional Experimental Analysis}

\subsection{Multi-Modal Backbone Analysis}
\label{appendix:model_analysis}

We evaluate MAG across seven state-of-the-art multi-modal foundation models to assess performance variation with backbone choice. Table~\ref{tab:model_appendix} reports results across all eight benchmarks.

\begin{table*}[h!t!b]
\centering
\caption{Performance of MAG with different multi-modal backbones (accuracy \%). Gemini-2.0-Flash provides the most balanced performance across task types.}
\label{tab:model_appendix}
\setlength{\tabcolsep}{2.5pt}
  \renewcommand{\arraystretch}{1.0}
  \resizebox{1\textwidth}{!}{
\begin{small}
\begin{sc}
\begin{tabular}{lcccccccc}
\toprule
Model & EmoSet & Emotion6 & TextOCR & MMStar & CLEVR & MatchingMI & GQA & OKVQA \\
\midrule
Gemini-2.0-Flash & 77.0 & 65.0 & 91.7 & 70.3 & 92.0 & 99.7 & 62.3 & 56.7 \\
Gemini-3.0-Flash & 84.7 & 59.3 & 55.7 & 92.3 & 96.0 & 87.3 & 68.0 & 57.3 \\
GPT-4o & 73.7 & 59.3 & 34.7 & 19.3 & 91.3 & 89.3 & 26.7 & 42.0 \\
GPT-4.1 & 76.0 & 60.7 & 50.3 & 33.3 & 93.7 & 79.3 & 37.3 & 52.0 \\
GPT-5 & 78.7 & 60.3 & 89.0 & 45.7 & 96.3 & 99.7 & 71.3 & 68.3 \\
Qwen-VL-Plus & 71.7 & 23.0 & 89.3 & 30.0 & 76.0 & 96.3 & 68.0 & 61.3 \\
Qwen-VL-Max & 73.7 & 60.0 & 79.7 & 50.0 & 97.3 & 90.0 & 71.0 & 64.0 \\
\bottomrule
\end{tabular}
\end{sc}
\end{small}
}
\end{table*}

\noindent
\textbf{Overall trends.}
We observe substantial performance variance across backbones, indicating that MAG is sensitive to the underlying multi-modal reasoning and perception capabilities. While all models benefit from MAG, their relative strengths differ significantly across task types.

\noindent
\textbf{Balanced vs. specialized performance.}
Gemini-2.0-Flash achieves the most balanced performance, maintaining consistently strong results across both perception-heavy tasks (e.g., TextOCR, CLEVR) and reasoning-intensive benchmarks (e.g., GQA, OKVQA). In contrast, Gemini-3.0-Flash exhibits more polarized behavior, achieving state-of-the-art results on MMStar and CLEVR, but suffering a notable drop on TextOCR, suggesting potential trade-offs between visual reasoning and text recognition.

\noindent
\textbf{Effect of backbone capability.}
Stronger general-purpose models such as GPT-5 demonstrate clear advantages on high-level reasoning tasks (e.g., GQA, OKVQA), while also maintaining competitive performance on structured reasoning benchmarks like CLEVR and MatchingMI. However, earlier generations (e.g., GPT-4o, GPT-4.1) show clear limitations on multi-modal understanding tasks such as MMStar and TextOCR, highlighting the importance of improved visual-text alignment in newer models.

\noindent
\textbf{Open-source vs. proprietary models.}
Qwen-VL variants exhibit competitive performance on several benchmarks, particularly MatchingMI and GQA, but show instability on tasks requiring fine-grained semantic understanding (e.g., Emotion6). Notably, Qwen-VL-Plus suffers a significant drop on Emotion6, indicating that emotion recognition remains challenging for some open-source models.

\noindent
\textbf{Key takeaway.}
These results suggest that MAG is robust across a wide range of backbones, but its effectiveness is amplified by models with strong cross-modal alignment and reasoning capabilities. In particular, backbones that achieve a better balance between perception and reasoning (e.g., Gemini-2.0-Flash, GPT-5) tend to yield the most consistent gains across diverse benchmarks.
\label{appendix:additional_analysis}

\subsection{Scalability Analysis}
\label{appendix:scalability}

We further analyze the scalability of our method in terms of both computational and memory complexity.

As shown in Eq.~(4) of the main paper, the steady-state solution of graph propagation is:
\[
I^{(m)}_{(\infty)} = (1-\alpha)(\mathbf{I}-\alpha \hat{W}^{(m)})^{-1} I^{(m)}_{(0)}.
\]

The main computational cost arises from solving the linear system involving $(\mathbf{I}-\alpha \hat{W}^{(m)})$. In practice, $\hat{W}^{(m)}$ is a sparse affinity matrix constructed via $k$-nearest neighbors. Therefore, this reduces to solving a \textbf{sparse linear system}, rather than performing dense matrix inversion.

Such systems can be efficiently solved using iterative methods (e.g., Conjugate Gradient, Lanczos) or sparse Cholesky decomposition, with complexity typically ranging from $O(n \log n)$ to $O(n^{1.5})$, significantly lower than the $O(n^3)$ complexity of dense inversion. Memory consumption scales linearly as $O(nk)$, since only non-zero edges are stored.

\paragraph{Empirical Validation.}
We validate the above analysis with empirical measurements shown in Table~\ref{tab:graph_scalability}.

\begin{table}[h]
\centering
\caption{Latency and memory usage of graph propagation.}
\label{tab:graph_scalability}
\begin{tabular}{lcccc}
\toprule
Sample Number & 600 & 800 & 1000 & 1200 \\
\midrule
Edge Number & 6000 & 8000 & 10000 & 12000 \\
Latency (ms) & 5.7 & 7.6 & 9.5 & 11.4 \\
Memory Usage (MB) & 12.8 & 20.1 & 28.0 & 38.5 \\
\bottomrule
\end{tabular}
\end{table}

As the number of samples increases, the number of edges grows linearly, consistent with the $O(nk)$ complexity. The latency increases approximately linearly (doubling the samples roughly doubles runtime), confirming that our method avoids cubic complexity. Memory usage also exhibits near-linear growth, with minor deviations due to implementation overhead.

Overall, both theoretical and empirical results demonstrate that our approach remains efficient and scalable for larger unlabeled datasets.

\subsection{Pseudo-Label Accuracy Analysis}
\label{appendix:pseudo_label}

We present a quantitative comparison between different pseudo-label selection strategies in Table~\ref{tab:pseudo_label}.

\begin{table}[h]
\centering
\caption{Pseudo-label accuracy (p-label acc) and final performance.}
\label{tab:pseudo_label}
\begin{tabular}{lccccc}
\toprule
Metric & EmoSet & Emotion6 & TextOCR & MMStar & GQA \\
\midrule
Random (p-label acc) & 0.667 & 0.556 & 0.833 & 0.756 & 0.489 \\
Ours (p-label acc) & 0.778 & 0.400 & 0.933 & 0.800 & 0.578 \\
\midrule
Random (final) & 0.733 & 0.657 & 0.870 & 0.687 & 0.570 \\
Ours (final) & 0.770 & 0.650 & 0.917 & 0.703 & 0.623 \\
\bottomrule
\end{tabular}
\end{table}

We observe a strong positive correlation between pseudo-label accuracy and final performance. In 4 out of 5 datasets, our method outperforms random selection in both pseudo-label quality and downstream performance, validating the effectiveness of our selection strategy.

\paragraph{Robustness to Real-World Noise.}
To address real-world challenges such as cross-modal inconsistencies, our method adopts modality-specific disentanglement. Each modality is processed independently, preventing noise in one modality (e.g., low-quality images) from contaminating the other during propagation.

As demonstrated in Section~4.6, the proposed late-fusion strategy consistently outperforms alternative designs across seven datasets. By selectively integrating reliable steady-state representations from each modality, our method exhibits strong robustness to semantic gaps and noisy inputs.

\clearpage
\newpage

\section{Qualitative Analysis}
\label{appendix:qualitative}

\subsection{Comparison with Baselines}

\begin{figure*}[htb]
\vskip 0.2in
\begin{center}
\includegraphics[width=0.95\textwidth]{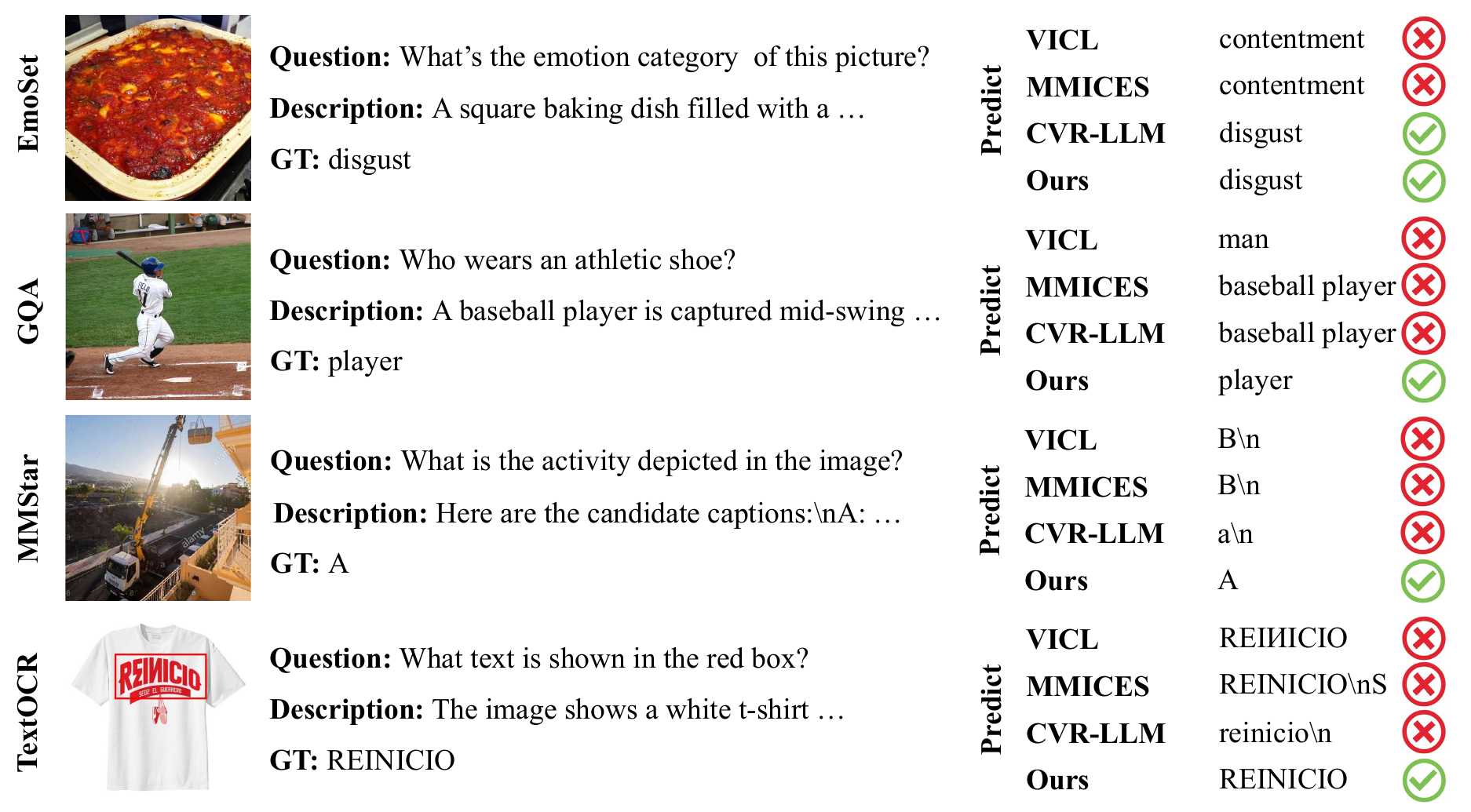}
\caption{Qualitative comparison between MAG and baselines (VICL, MMICES, CVR-LLM) across four tasks. Our method consistently produces more accurate predictions by selecting semantically relevant demonstrations via graph-based propagation.}
\label{fig:qualitative_baseline}
\end{center}
\vskip -0.2in
\end{figure*}

Figure~\ref{fig:qualitative_baseline} compares MAG with strong baselines across emotion recognition (EmoSet), visual QA (GQA), reasoning (MMStar), and OCR (TextOCR). Our method consistently produces more accurate predictions. In emotion recognition, we correctly identify subtle affective cues (\emph{disgust}) while baselines produce generic interpretations. For reasoning tasks, graph-based selection helps distinguish linguistically plausible but visually inconsistent options. In OCR, we accurately recognize scene text (\emph{REINICIO}) where baselines hallucinate. These examples demonstrate that multi-modal graph propagation enables selection of globally relevant demonstrations beyond local similarity.

\subsection{Ablation Visualizations}

\begin{figure*}[t]
\begin{center}
\includegraphics[width=0.95\textwidth]{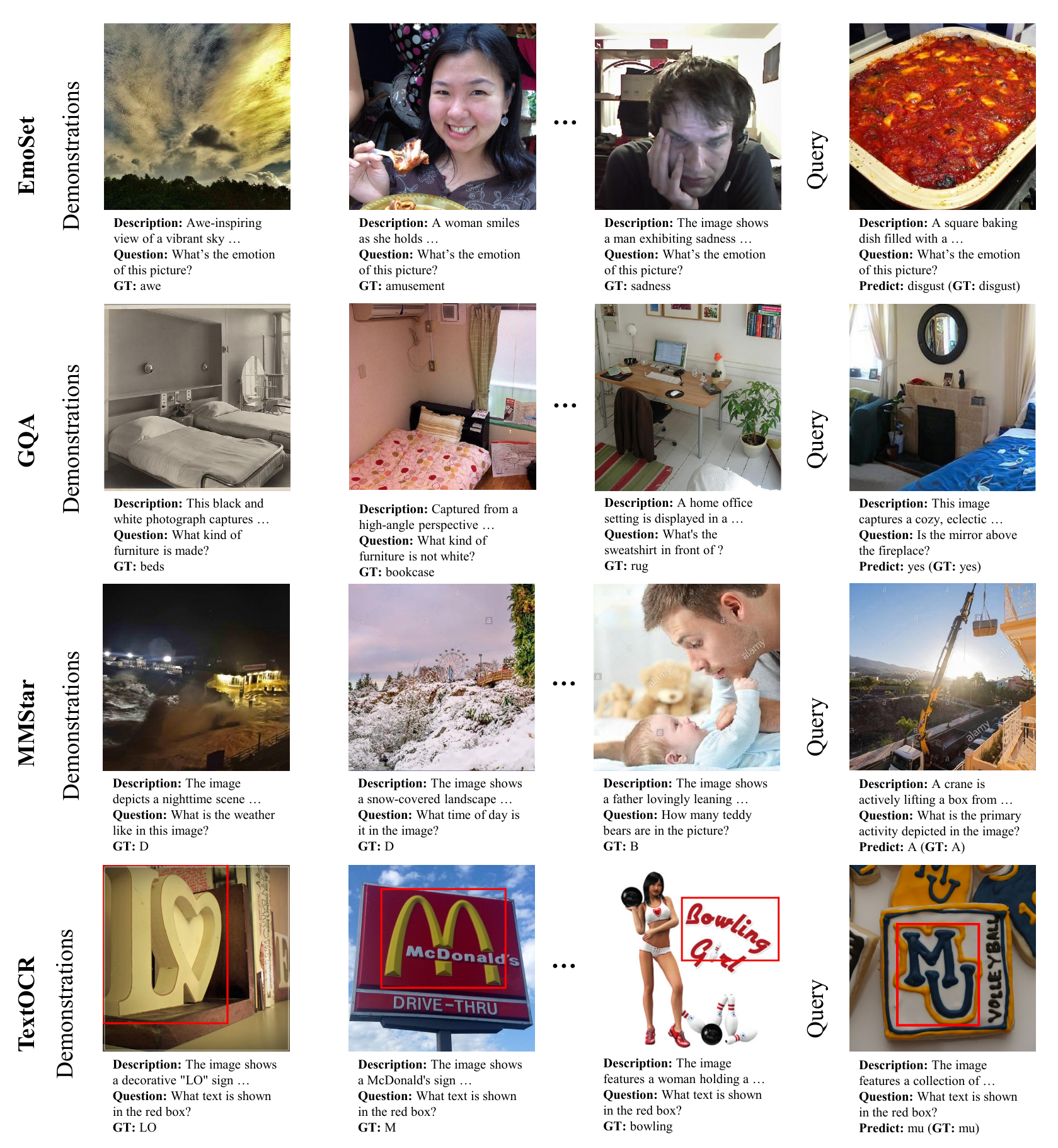}
\caption{Visualization of the full MAG method across EmoSet, GQA, MMStar, and OCR tasks.}
\label{fig:qualitative_ablation_w}
\end{center}
\end{figure*}

\begin{figure*}[t]
\begin{center}
\includegraphics[width=0.95\textwidth]{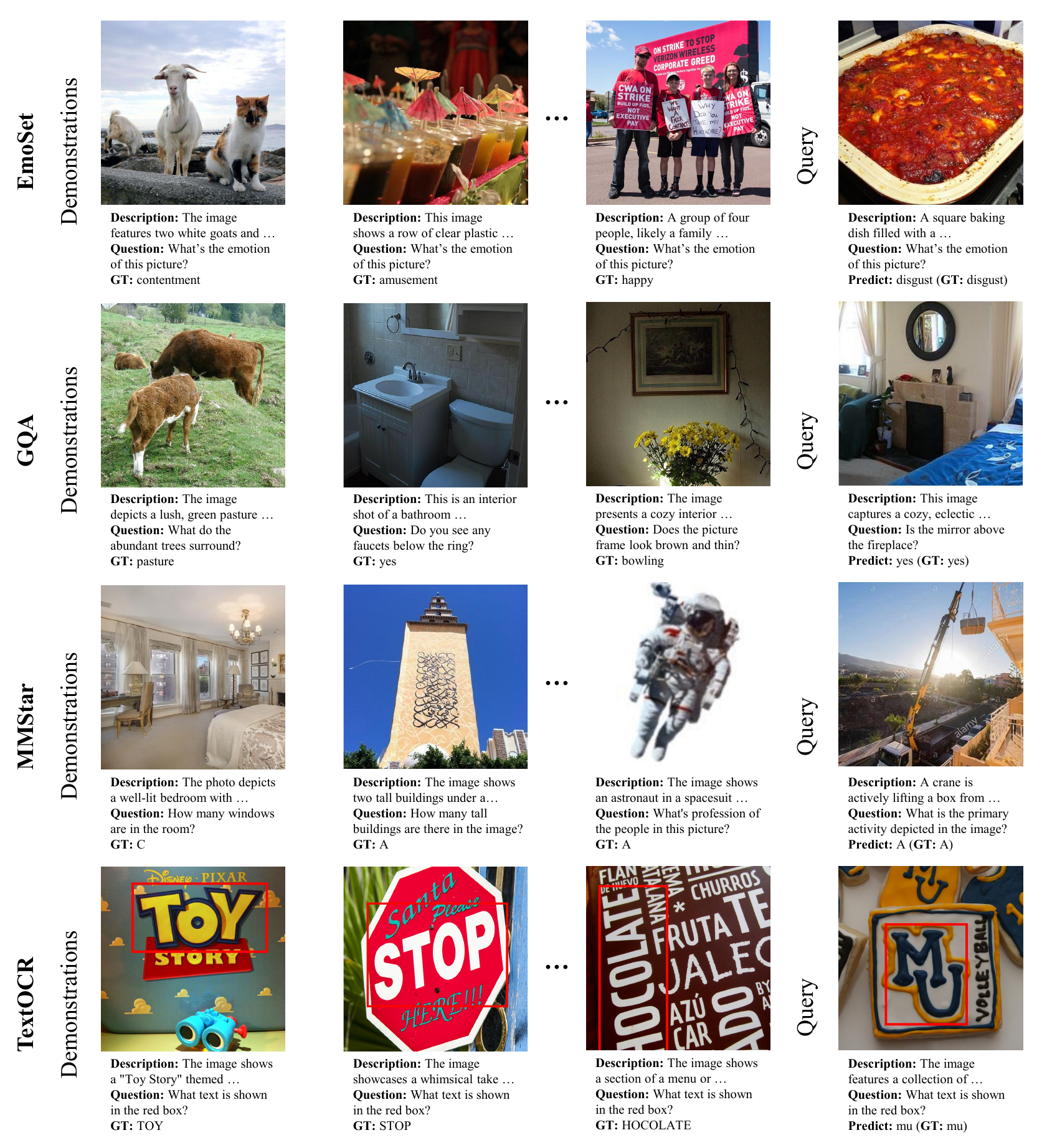}
\caption{MAG with random Stage 1 pseudo-label selection (vs. relevance-guided). Random selection leads to weaker semantic alignment and less accurate predictions.}
\label{fig:qualitative_ablation_1}
\end{center}
\end{figure*}

\begin{figure*}[t]
\begin{center}
\includegraphics[width=0.95\textwidth]{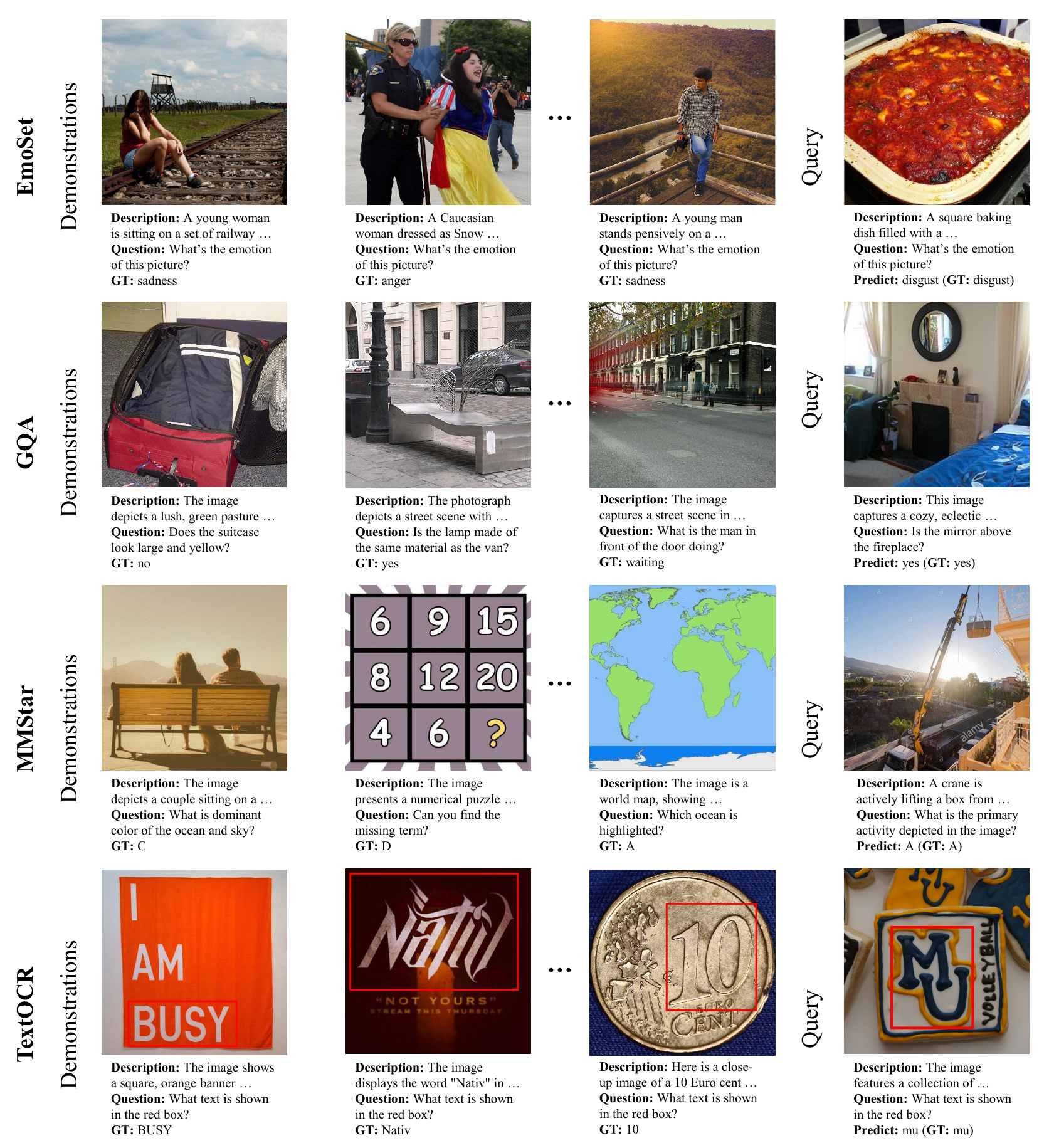}
\caption{MAG with random Stage 2 demonstration selection (vs. graph-based). Random selection introduces noisy demonstrations, harming reasoning and OCR tasks.}
\label{fig:qualitative_ablation_2}
\end{center}
\end{figure*}

Figures~\ref{fig:qualitative_ablation_w}--\ref{fig:qualitative_ablation_2} present qualitative ablations comparing the full method with randomized Stage 1 (pseudo-label selection) and randomized Stage 2 (demonstration selection).

When Stage 1 selection is randomized (Figure~\ref{fig:qualitative_ablation_1}), the retrieved demonstrations are less semantically aligned with the queries, which in turn yields less stable emotion recognition and weaker object-centric reasoning. Moreover, random selection increases the incidence of corrupted OCR text outputs, suggesting that the model is less likely to incorporate informative unlabeled samples under this setting.

Replacing Stage 2 with random demonstration selection (Figure~\ref{fig:qualitative_ablation_2}) further degrades performance. Irrelevant demonstrations introduce noisy visual-text associations, harming multi-choice reasoning and scene-text understanding.

In contrast, the full model (Figure~\ref{fig:qualitative_ablation_w}) consistently selects semantically coherent demonstrations, producing accurate predictions across all task types. These qualitative results verify that both relevance-guided pseudo-labeling (Stage 1) and graph-based demonstration selection (Stage 2) are essential for robust multi-modal reasoning.

\clearpage
\newpage

\section{Evaluation Protocol}
\label{appendix:evaluation_protocol}

Following prior multi-modal in-context learning works, we adopt an exact containment-based evaluation criterion for all benchmarks. Specifically, given the model response $r$ and the ground-truth answer $y$, a prediction is considered correct if the ground-truth text appears in the generated response:

\[
\mathrm{Correct}(r, y)=
\begin{cases}
1, & \text{if } y \subseteq r, \\
0, & \text{otherwise}.
\end{cases}
\]

That is, if the response contains the ground-truth answer as a substring, the prediction is counted as correct; otherwise, it is treated as incorrect. We report the average accuracy over all evaluation samples.
This evaluation protocol is particularly suitable for open-ended multi-modal generation settings, where models may produce additional explanatory text while still containing the correct answer.

\section{Prompts for ICL}
\label{appendix:prompts}

The prompts we use for in-context learning on different tasks are as follows. 

\begin{itemize}
\item For visual emotion recognition task(EmoSet and Emotion6), the prompt we use in our method is "What is the emotion category of this image? Give only the emotion category, and no extra commentary, formatting, or chattiness. You can only make prediction from the following categories: [{Label List}]. Given an image and its description, please predict the emotion category of this image. Here are several examples. Image: {image-1}, Description: {description-1}, Question: {question-1}, Emotion category: {label-1}. Image: {image-2}, Description: {description-2}, Question: {question-2}, Emotion category: {label-2}... Image: {image-N}, Description: {description-N}, Question: {question-N}, Emotion category: {label-N}. Image: {image-q}, Description: {description-q}, Question: {question-q}, Emotion category:".   

\item For TextOCR, MMStar, CLEVR, GQA and OKVQA, the prompt is "Given an image and its description, please answer a question about the image. Here are several examples. Image: {image-1}, Description: {description-1}, Question: {question-1}, Answer: {label-1}. Image: {image-2}, Description: {description-2}, Question: {question-2}, Answer: {label-2}... Image: {image-N}, Description: {description-N}, Question: {question-N}, Answer: {label-N}. Image: {image-q}, Description: {description-q}, Question: {question-q}, Answer:". 

\item For MatchingMI, the prompt is "Given two images and their description, please answer a question about the two images. Here are several examples. Images: {image-1-A}{image-1-B}, Description: {description-1}, Question: {question-1}, Answer: {label-1}. Image: {image-2-A}{image-2-B}, Description: {description-2}, Question: {question-2}, Answer: {label-2}... Image: {image-N-A}{image-N-B}, Description: {description-N}, Question: {question-N}, Answer: {label-N}. Image: {image-q-A}{image-q-B}, Description: {description-q}, Question: {question-q}, Answer:".
\end{itemize}


\end{document}